\documentclass{article}

\usepackage{ijcai16}

\usepackage{times}

\usepackage{amssymb}
\usepackage{amsmath}

\usepackage{amsthm}
\usepackage{bbm}
\usepackage{graphicx}
\usepackage{subfigure}
\usepackage{url}
\usepackage{threeparttable}
\usepackage{multirow}
\usepackage[table,xcdraw]{xcolor}
\usepackage[linesnumbered,vlined,ruled]{algorithm2e}
\usepackage{tabularx, booktabs}

\graphicspath{{figures/}}

\theoremstyle{plain}

\title{Overview: Generalizations of Multi-Agent Path Finding to Real-World Scenarios}

\author{
Hang Ma, Sven Koenig, Nora Ayanian, Liron Cohen\\{\Large\bf Wolfgang H\"onig, T.~K. Satish Kumar, Tansel Uras, Hong Xu}\\
Department of Computer Science\\ University of Southern California\\
\{hangma,skoenig,ayanian,lironcoh,whoenig,turas,hongx\}@usc.edu, tkskwork@gmail.com
\AND
  Craig Tovey\\
  School of ISyE\\
  Georgia Institute of Technology\\
  ctovey@isye.gatech.edu \And
  Guni Sharon\\
  Department of Computer Science\\
  The University of Texas at Austin\\
  gunisharon@gmail.com
}

\begin{document}

\maketitle

\begin{abstract}
  Multi-agent path finding (MAPF) is well-studied in artificial intelligence,
  robotics, theoretical computer science and operations research. We discuss issues that arise when generalizing MAPF methods to real-world scenarios and four research directions that address them. We emphasize the importance of addressing these issues as opposed to developing faster methods for the standard formulation of the MAPF problem.
\end{abstract}

\section{Introduction}
Multi-agent path finding (MAPF) has been well-studied by researchers from
artificial intelligence, robotics, theoretical computer science and operations
research. The task of (standard) MAPF is to find the paths for multiple agents
in a given graph from their current vertices to their targets without colliding with other agents, while at the same time optimizing a cost function. Existing MAPF methods use, for example, reductions to problems from satisfiability, integer linear programming or answer set programming~\cite{YuLav13ICRA,erdem2013general,Surynek15} or optimal, bounded-suboptimal or suboptimal search methods~\cite{WHCA,WHCA06,Ryan08,WangB08,ODA,ODA11,WangB11,PushAndSwap,DBLP:journals/ai/SharonSGF13,PushAndRotate,ECBS,EPEJAIR,MStar,ICBS,DBLP:journals/ai/SharonSFS15}.

\begin{figure}
\label{kiva_warehouse}
\center
  \includegraphics[height=62pt]{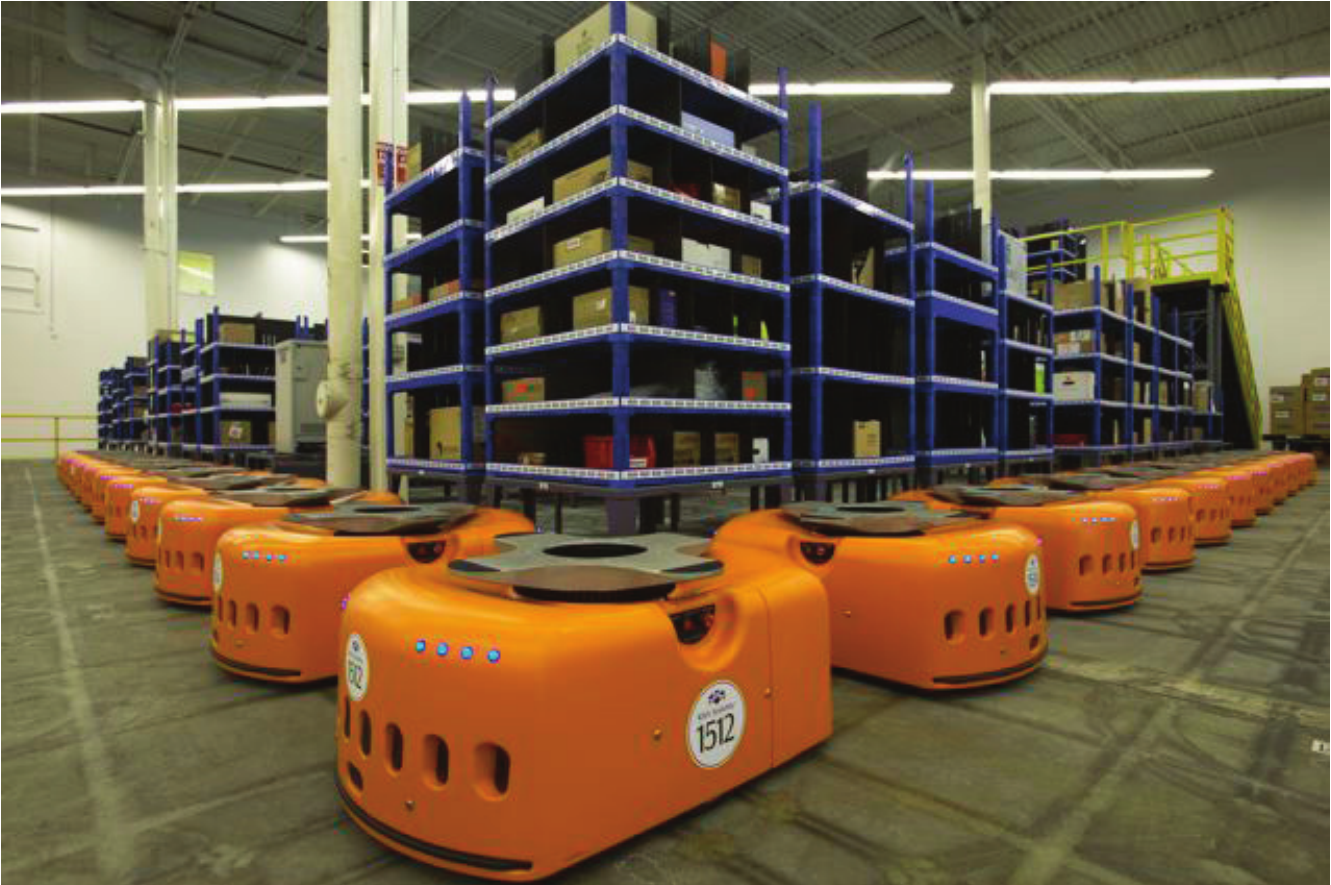}\includegraphics[height=62pt]{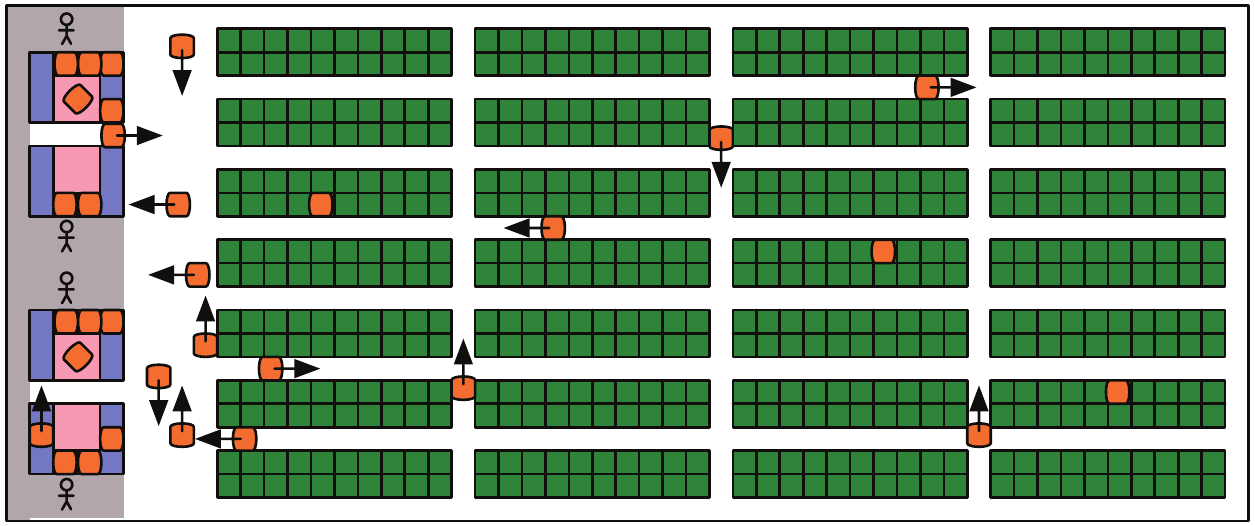}
  \caption{Autonomous drive units and storage pods that contain products and can be moved by the drive units (left) and the layout of a typical Kiva warehouse system (right)~\protect\cite{kiva}.}
\end{figure}

We have recently studied various issues that arise when generalizing MAPF to
real-world scenarios, including Kiva (Amazon Robotics) warehouse
systems~\cite{kiva} (Figure~\ref{kiva_warehouse}) and autonomous aircraft
towing vehicles~\cite{airporttug16}. These issues can be categorized into two
general concerns: 1. Developing faster methods for the standard formulation of
the MAPF problem is insufficient because, in many real-world scenarios, new
structure can be exploited or new problem formulations are
required. 2. Studying MAPF or its new formulations only as combinatorial
optimization problems is insufficient because the resulting MAPF solutions also need to be executed. We discuss four research directions that address both concerns from different perspectives:

\begin{enumerate}
  \item In many real-world multi-agent systems, agents are partitioned into
    teams, targets are given to teams, and each agent in a team needs
    to get assigned a target from the team, before one finds paths for all agents. We have formulated the \emph{combined target assignment and path finding} (TAPF) problem for teams of agents to address this issue. We have also developed an optimal TAPF method that scales to dozens of teams and hundreds of agents~\cite{MaAAMAS16}.
  \item In many real-world multi-agent systems, agents are anonymous
    (exchangeable), but their payloads are non-anonymous (non-exchangeable)
    and need to be delivered to given targets. The agents can often exchange their payloads in such systems. We have formulated the \emph{package-exchange robot routing} (PERR) problem as a first attempt to tackle more general transportation problems where payload transfers are allowed~\cite{MaAAAI16}. In this context, we have also proved the hardness of approximating optimal MAPF solutions.
  \item In many real-world multi-agent systems, the 
    consistency of agent motions and the resulting predictability of agent
    motions is important (especially in work spaces
    shared by humans and agents), which is not taken into account by existing
    MAPF methods. We have exploited the problem structure of given MAPF
    instances in two stages: In the first stage, we have developed a scheme
    for finding paths for the agents that include many edges from
    user-provided highways, which achieves consistency and predictability of agent motions~\cite{DBLP:conf/socs/CohenUK15}. In the second stage, we have developed methods that automatically generate highways~\cite{CohenUK16}.
  \item MAPF is mostly motivated by navigation or motion planning for
    multi-robot systems. However, the optimality or bounded-suboptimality of
    MAPF solutions does not necessarily entail their robustness, especially
    given the imperfect plan-execution capabilities of real-world robots. We have developed a framework that efficiently postprocesses the output of a MAPF method to create a plan-execution schedule that can be executed by real-world multi-robot systems~\cite{HoenigICAPS16}.
\end{enumerate}

We now showcase the practicality of these research directions to demonstrate that, in order to generalize MAPF methods to real-world scenarios, addressing both concerns is as important, if not more, than developing faster methods for the standard formulation of the MAPF problem.

\section{Combined Target Assignment and Path Finding (TAPF) for Teams of Agents}

Targets are often given to teams of agents. Each agent in a team
needs to get assigned a target given to its team so that the paths of the agents from their current vertices to their target optimize a cost function. For example, in a Kiva warehouse system, the drive units that relocate storage pods from the inventory stations to the storage locations form a team because each of them needs to get assigned an available storage location. Previous MAPF methods assume that each agent is assigned a target in advance by some target-assignment procedure but, to achieve optimality, we have formulated TAPF, which couples the target-assignment and the path-finding problems and defines one common objective for both of them. In TAPF, the agents are partitioned into teams. Each team is given the same number of unique targets as there are agents in the team. The task of TAPF is to assign the targets to the agents and plan collision-free paths for the agents from their current vertices to their targets in a way such that each agent moves to exactly one target given to its team, all targets are visited and the makespan (the earliest time step when all agents have reached their targets and stop moving) is minimized. Any agent in a team can get assigned a target of the team, and the agents in the same team are thus exchangeable. However, agents in different teams are not exchangeable. TAPF can be viewed as a generalization of (standard) MAPF and an anonymous variant of MAPF:

\begin{itemize}
  \item \textbf{(Standard) MAPF} results from TAPF if every team consists of exactly one agent and the number of teams thus equals the number of agents. The assignments of targets to agents are pre-determined and the agents are thus non-anonymous (non-exchangeable).
  \item \textbf{The anonymous variant of MAPF} (also called goal-invariant
    MAPF) results from TAPF if only one team exists (that consists of all
    agents). The agents can get assigned any target and are thus
    exchangeable. It can be solved optimally in polynomial time with
    flow-based MAPF methods~\cite{YuLav13STAR,MTurpin2014}.
\end{itemize}

The state-of-the-art optimal TAPF method, called the Conflict-Based Min-Cost
Flow~\cite{MaAAMAS16}, combines search and flow-based MAPF methods. It
generalizes to dozens of teams and hundreds of agents.

\section{Package-Exchange Robot Routing (PERR) and New Complexity Results for MAPF}

Agents are often anonymous but carry payloads (packages) that are assigned
targets and are thus non-anonymous. For example, in a Kiva warehouse system,
the drive units are anonymous but the storage pods they carry are assigned
storage locations and are thus non-anonymous. If each agent carries one package, the problem is equivalent to (standard) MAPF. In reality, the packages can often be transferred among agents, which results in more general transportation problems, for example, ride-sharing with passenger transfers~\cite{DBLP:conf/aaai/ColtinV14} and package delivery with robots in offices~\cite{DBLP:conf/ijcai/VelosoBCR15}. We have formulated PERR as a first step toward understanding these problems~\cite{MaAAAI16}. In PERR, each agent carries one package, any two agents in adjacent vertices can exchange their packages, and each package needs to be delivered to a given target. PERR can thus be viewed as a modification of (standard) MAPF:

\begin{itemize}
  \item Packages in PERR can be viewed as agents in (standard) MAPF which move by themselves.
  \item Two packages in adjacent vertices are allowed to exchange their vertices in PERR but two agents in adjacent vertices are not allowed to exchange their vertices in (standard) MAPF.
\end{itemize}

$K$-PERR is a generalization of PERR where packages are partitioned into $K$
types and packages of the same type are exchangeable. Since, in TAPF, agents
are partitioned into teams and agents in the same team are exchangeable,
$K$-PERR can be viewed as a modification of TAPF with $K$ teams in the same
sense that PERR can be viewed as a modification of (standard) MAPF. We have
proved the hardness of approximating optimal PERR and $K$-PERR solutions (for
$K\ge2$). One corollary of our study is that both MAPF and TAPF are NP-hard to
approximate within any factor less than 4/3 for makespan minimization, even
when there are only two teams for TAPF. We have also demonstrated that the
addition of exchange operations to MAPF does not reduce its complexity
theoretically but makes PERR easier to solve than MAPF experimentally. There
is a continuum of problems that arise in different real-world scenarios: ``One
agent with many packages'' yields the classic rural postman problem; ``as many
agents as packages'' yields MAPF, TAPF or PERR. Understanding both extremes
helps one to attack the middle, as required by many other real-world
scenarios.

\section{Exploitation of Problem Structure and Predictability of Motions}

\begin{figure}
\center
  \includegraphics[width=\columnwidth]{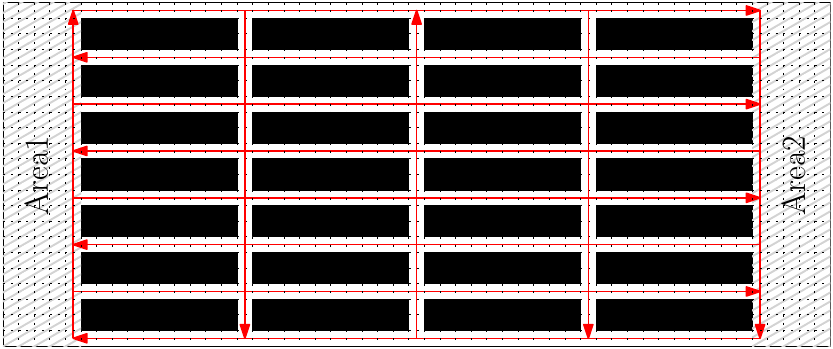}
  \caption{User-provided highways in a simulated Kiva warehouse system.}\label{kiva_arrows}
\end{figure}

Agents share their work spaces with humans, and the consistency of their
motions and the resulting predictability of their motions are important for the safety of the humans, which existing MAPF methods do not take into account. This motivates us to exploit the problem structure of given MAPF instances and design a scheme that encourages agents to move along user-provided sets of edges (called highways)~\cite{DBLP:conf/socs/CohenUK15}. We use highways in the context of a simple inflation scheme based on the ideas behind
experience graphs~\cite{PhillipsCCL12} to derive new heuristic values that
encourage MAPF methods to return paths that include the edges of the highways,
which avoids head-to-head collisions among agents and achieves 
consistency and predictability of their motions. For example, in a Kiva warehouse system, we can
design highways along the narrow passageways between the storage locations as
shown by the arrows in Figure~\ref{kiva_arrows}. We have demonstrated in a
simulated Kiva warehouse system that such highways accelerate MAPF methods
significantly while maintaining the desired bounded-suboptimality of the MAPF solution costs. The problem structure of TAPF and PERR instances can also be exploited with the same methods. We have also developed methods that automatically generate highways that are competitive with user-provided ones in our feasibility study~\cite{CohenUK16}.

\section{Dealing with Imperfect Plan-Execution Capabilities}

State-of-the-art MAPF or TAPF methods can find collision-free paths for
hundreds of agents optimally or with user-provided sub-optimality guarantees
in a reasonable amount of computation time. They perform well even in
cluttered and tight environments, such as Kiva warehouse systems. However, agents often have imperfect plan-execution capabilities and are not able to synchronize their motions perfectly, which can result in frequent and time-expensive replanning. Therefore, we have proposed a framework that makes use of a simple temporal network to postprocess a MAPF solution efficiently to create a plan-execution schedule that works for non-holonomic robots, takes their maximum translational and rotational velocities into account, provides a guaranteed safety distance between them and exploits slack (defined as the difference of the latest and earliest entry times of locations) to absorb imperfect plan executions and avoid time-intensive replanning in many cases~\cite{HoenigICAPS16}. This framework has been evaluated in simulation and on real robots. TAPF and PERR methods can also be applied in the same framework. Issues to be addressed in future work include adding user-provided safety distances, additional kinematic constraints, planning with uncertainty and replanning.

\section{Conclusions}
We discussed four research directions that address issues that arise when generalizing MAPF methods to real-world scenarios and exploit either the problem structure or existing MAPF methods. Our goal was to point out interesting research directions for researchers working in the field of MAPF.

\section{Acknowledgments}

The research at USC was supported by ARL under grant number W911NF-14-D-0005, ONR under grant numbers N00014-14-1-0734 and N00014-09-1-1031, NASA via Stinger Ghaffarian Technologies and NSF under grant numbers 1409987 and 1319966. The views and conclusions contained in this document are those of the authors and should not be interpreted as representing the official policies, either expressed or implied, of the sponsoring organizations, agencies or the U.S. government.

\bibliographystyle{named}
\bibliography{references}

\end{document}